%% file: main.tex
\documentclass[nohyperref]{article}

\usepackage{microtype}
\usepackage{graphicx}
\usepackage{subfigure}
\usepackage{booktabs} 

\usepackage{hyperref}
\usepackage{enumitem}

\usepackage[accepted]{icml2022}

\usepackage{amsmath}
\usepackage{amssymb}
\usepackage{mathtools}
\usepackage{amsthm}
\usepackage{multirow}

\usepackage[capitalize,noabbrev]{cleveref}

\theoremstyle{plain}

\theoremstyle{definition}

\theoremstyle{remark}

\newcommand{\yx}[1]{\textcolor{red}{[yx: #1]}}
\newcommand\T{\rule{0pt}{2.6ex}} 

\usepackage[textsize=tiny]{todonotes}
\usepackage{picinpar}

\icmltitlerunning{Neural Implicit 3D Shapes from Single Images with Spatial Patterns}

\begin{document}

\twocolumn[
\icmltitle{Neural Implicit 3D Shapes from Single Images with Spatial Patterns}




\begin{icmlauthorlist}
\icmlauthor{Yixin Zhuang}{to}
\icmlauthor{Yunzhe Liu}{to}
\icmlauthor{Yujie Wang}{to,goo}
\icmlauthor{Baoquan Chen}{to}
\end{icmlauthorlist}

\icmlaffiliation{to}{Peking University}
\icmlaffiliation{goo}{Shandong University}

\icmlcorrespondingauthor{Yixin Zhuang}{yixin.zhuang@gmail.com}
\icmlcorrespondingauthor{Baoquan Chen}{baoquan.chen@gmail.com}

\icmlkeywords{Machine Learning, ICML}

\vskip 0.3in
]



\printAffiliationsAndNotice{}  

\begin{abstract}

Neural implicit functions have achieved impressive results for reconstructing 3D shapes from single images. However, the image features for describing 3D point samplings of implicit functions are less effective when significant variations of occlusions, views, and appearances exist from the image. To better encode image features, we study a geometry-aware convolutional kernel to leverage geometric relationships of point samplings by the proposed \emph{spatial pattern}, i.e., a structured point set. Specifically, the kernel operates at 2D projections of 3D points from the spatial pattern. Supported by the spatial pattern, the 2D kernel encodes geometric information that is crucial for 3D reconstruction tasks, while traditional ones mainly consider appearance information. Furthermore, to enable the network to discover more adaptive spatial patterns for further capturing non-local contextual information, the kernel is devised to be deformable manipulated by a spatial pattern generator. Experimental results on both synthetic and real datasets demonstrate the superiority of the proposed method.
Pre-trained models, codes, and data are available at \href{https://github.com/yixin26/SVR-SP}{https://github.com/yixin26/SVR-SP}.

\end{abstract}

\input{intro}

\input{related}

\input{method}

\input{results}

\input{conclusion}


\input{bbl}
\input{appendix}

\end{document}

%% file: intro.tex
\section{Introduction}
3D shape reconstruction from a single image has been one of the central problems in computer vision.
Empowering the machines with the ability to perceive the imagery and infer the underlying 3D shapes can benefit various downstream tasks, such as augmented reality, robot navigation, etc.
However, the problem is overly ambiguous and ill-posed, and thus remains highly challenging, due to the information loss and occlusion occurred during the imagery capture.

In recent years, many deep learning based methods have been proposed to infer 3D shapes from single images.
These methods rely on learning shape priors from a large amount of shape collections, for reasoning the underlying shape of unseen images.
To this end, various learning frameworks have been proposed that exploit different 3D shape representations, including point sets~\cite{fan2017point,Achlioptas2018}, voxels~\cite{wu2016learning,wu2018learning}, polygonal meshes~\cite{groueix2018papier,wang2018pixel2mesh}, 
and implicit fields~\cite{OccNet,park2019deepsdf,chen2018implicit_decoder}.
In particular, implicit field-based models have shown impressive performance compared to the others.

Implicit field-based networks take a set of 3D samplings as input, and predict corresponding values in accordance with varying representations (e.g., occupancy, signed distance, etc.). Once the network is trained, 3D shapes are identified as the zero level of the predicted scalar fields using meshing methods such as Marching Cubes~\cite{lorensen1987marching}.
By conditioning the 3D shape generation on the extracted global feature of input image~\cite{OccNet,chen2018implicit_decoder}, the implicit networks are well-suited to reconstruct 3D shapes from single images.
However, this trivial combination often fails to reconstruct fine geometric details and produces overly smoothed surfaces.

\begin{figure*}
\begin{center}
{\includegraphics[width=0.99\linewidth]{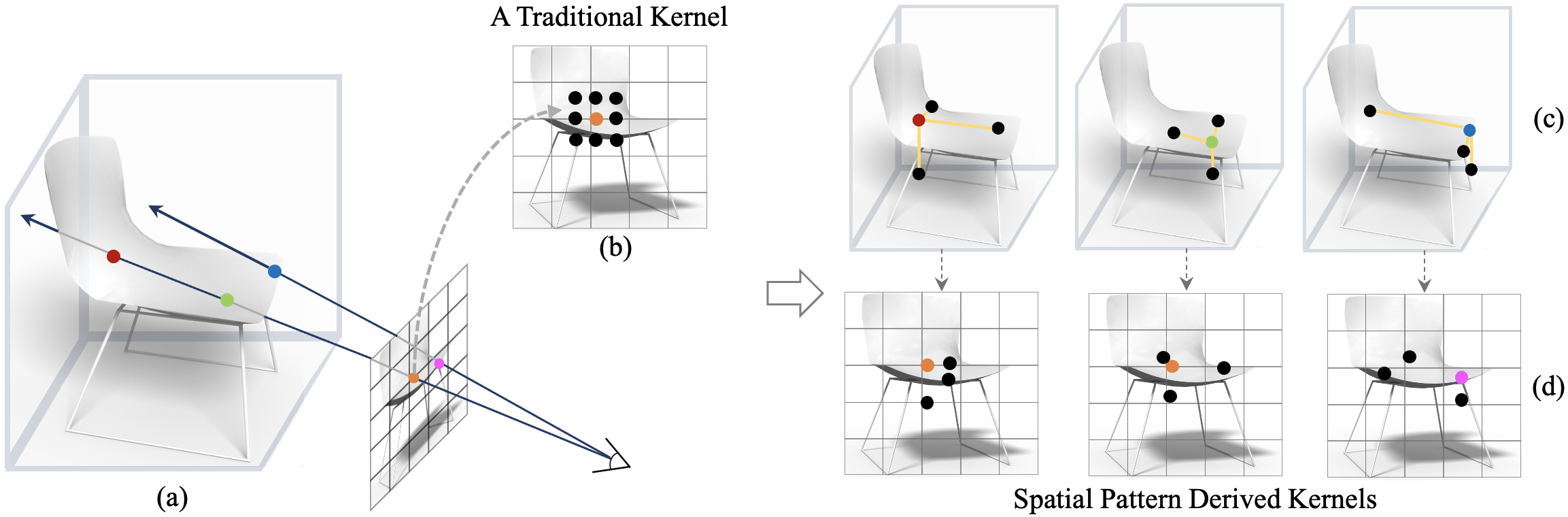} }
\end{center}
\caption{
Illustration of the pipeline of spatial pattern guided kernel. (a) shows that each 3D point sampling (colored differently) of the depicted shape is aligned to a 2D pixel by the given camera pose. 
{Compared with a 2D convolution kernel (b) that only considers neighbors located within a 2D regular local patch, the kernels in (d) derived from the proposed spatial patterns (c) explicitly exploit the underlying geometric relations for each pixel. 
As a result, the kernels in (d) encode the local image features for capturing both image contextual information and point geometry relations.}
}
\label{fig:teaser}
\end{figure*}

To address this issue, DISN~\cite{DISN} proposes a pixel-aligned implicit surface network where individual point sampling is conditioned on a learned local image feature obtained by projecting the point to the image plane according to the camera pose. With local image features, the network predicts a residual field for refinement.
However, the strategy of associating 3D samplings with learned local image features would become less effective when samplings are occluded from the observation view. 
Hence, to represent each point sampling with meaningful local image feature,  Ladybird~\cite{xu2020ladybird} utilizes the feature extracted from the 2D projection of its symmetric point obtained from the self-reflective symmetry of the object. The reconstruction quality is significantly improved upon DISN. 
%
Nevertheless, the strategy used in Ladybird is not sufficiently generic as the feature probably would have no intuitive meaning in the situation where the symmetric points are non-visible or the symmetry assumption does not hold.

As evidenced by DISN and Ladybird, the use of local image features is effective. 
In this paper, we introduce spatial pattern, a point-based geometric structure, to achieve better exploitation of local image features. 
The spatial pattern may include geometric relationships, e.g., symmetric, co-planar, and other structures that are less intuitive.
With the spatial pattern, a kernel operating in image space is derived to encode local image features of 3D point samplings.
Specifically, the pattern is formed by a fixed number of affinities around a 3D sampling, for which the corresponding 2D projections are utilized as the operation positions of the kernel. 
Although a traditional 2D convolution is possible to encode contextual information for the central point, it ignores the underlying geometric relations in original 3D space between pixels and encounters the limitations brought by the regular local area. A 2D deformable kernel \cite{dai2017deformable} is able to operate on irregular neighborhoods, but it is still not able to explicitly consider the underlying 3D geometric relations, which are important in 3D reconstruction tasks.

Figure~\ref{fig:teaser} shows the pipeline of the 3D spatial pattern guided 2D kernel.
As shown in Figure \ref{fig:teaser} (c-d), the kernels operate on points determined by spatial patterns for different point samplings. 
Specifically, the proposed kernel finds kernel points adaptively for each pixel, which consider its geometric-related positions ({e.g.}, symmetry locations) in the underlying 3D space, rather than only relying upon the appearance information. Furthermore, the spatial pattern is devised to be deformable to enable the network to discover more adaptive geometric relations for point samplings. In the experiments section, we analyze the learned 3D spatial pattern with visualization and statistics.
%

%

To demonstrate the effectiveness of spatial pattern guided kernel, we integrate it into a network based on a deep implicit network~\cite{DISN},
and extensively evaluate our model on the large collection of 3D shapes -- the ShapeNet Core dataset~\cite{ShapeNet2015} and Pix3D dataset~\cite{pix3d}. 
The experiments show that our method can produce state-of-the-art 3D shape reconstruction results from single images compared to previous works.
Ablation experiments and analyses are conducted to show the performance of different spatial pattern variants and the importance of individual points within the spatial pattern.

In this work, we make the following contributions.
\begin{itemize}[itemsep=2pt,topsep=0pt,parsep=2pt]
\item We present spatial pattern to provide the network with more flexibility to discover meaningful image features that explicitly consider the geometric relationships. 
\item We extend 2D deformable convolutional kernel with a 3D spatial pattern generator to discover meaningful geometric structures while encoding image features.
\end{itemize}

%% file: related.tex
\begin{figure*}[th!]
\begin{center}
{\includegraphics[width=0.99\linewidth]{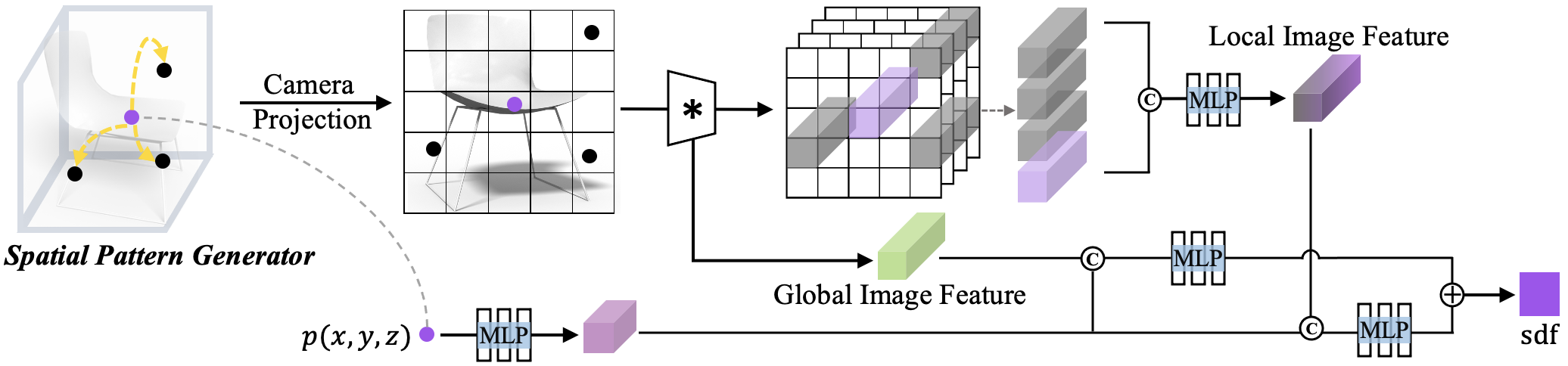} }
\end{center}
\caption{
The overview of our method. Given an image, our network predicts the signed distance field (SDF) for the underlying 3D object. To predict the SDF value for each point $p$, except for utilizing the global feature encoded from the image and the point feature directly inferred from $p$, local image features are fully exploited. Particularly, the local feature of a 3D point is encoded with a kernel in the image space whose kernel points are derived from a spatial pattern. *, $\copyright$ and $\oplus$ denote convolution, concatenation and sum operations respectively.
}
\label{fig:network}
\end{figure*}

\section{Related Work}

\paragraph{Deep Neural Networks for SVR.}

There has been a lot of research on single image reconstruction task. Recent works involve 3D representation learning, including points~\cite{fan2017point,lin2018learning,mandikal20183d}, voxels~\cite{choy20163d,wu2018learning,xie2019pix2vox}, meshes~\cite{groueix2018papier,wang2018pixel2mesh,wang20193dn,gkioxari2019mesh}
and primitives~\cite{niu2018im2struct,tang2019skeleton,wu2020pq}.
The representation can also be learned without knowing the underlying ground truth 3D shapes~\cite{kato2018neural,liu2019soft,liu2020neural,yan2016perspective,insafutdinov2018unsupervised,lin2018learning}.

In this line of research, AtlasNet~\cite{groueix2018papier} represents 3D shapes as the union of several surface elements that are generated from the learned multilayer perceptrons (MLPs).
Pixel2Mesh~\cite{wang2018pixel2mesh} generates genus-zero shapes as the deformations of ellipsoid template meshes. The mesh is progressively refined with higher resolutions using a graph convolutional neural network conditioned on the multi-scale image features.
3DN~\cite{wang20193dn} also deforms a template mesh to the target, trained with a differentiable mesh sampling operator pushing sampled points to the target position.

\paragraph{Implicit Neural Networks for SVR.}

The explicit 3D representations are usually limited by fixed shape resolution or topology. Alternatively, implicit functions for 3D objects have shown the advantages at representing complicated geometry ~\cite{chen2018implicit_decoder,DISN,xu2020ladybird,li2020d,niemeyer2020differentiable,liu2020dist,jiang2020sdfdiff,saito2019pifu}. 
ImNet~\cite{chen2018implicit_decoder} uses an MLP-based neural network to approximate the signed distance field (SDF) of 3D shapes and shows improved results in contrast to the explicit surface representations.
OccNet~\cite{OccNet} generates an implicit volumetric shape by inferring the probability of each grid cell being occupied or not. The shape resolution is refined by repeatedly subdividing the interest cells. 
While those methods are capable of capturing the global shape structure, the geometric details are usually missing. 
In addition to the holistic shape description, DISN~\cite{DISN} adds a local image feature for each 3D point computed by aligning the image to the 3D shape using an estimated camera pose.
With global and local features, DISN recovers much better geometric details and outperforms state-of-the-art methods.
The local image feature of each 3D point sampling can be further augmented with its self-symmetry point in the situation of self-occlusion, as shown in Ladybird~\cite{xu2020ladybird}. Compared to Ladybird, we investigate a more general point structure, the spatial pattern, along with a deformable 2D kernel derived from the pattern, to encode geometric relationships for local image features.

\paragraph{Deformable Convolutional Networks.}
Deformable convolution predicts a dynamic convolutional filter for each feature position~\cite{dai2017deformable}. Compared to locally connected convolutions, deformable convolution enables the exploration of non-local contextual information.  
The idea was originally proposed for image processing and then extended for learning features from natural language ~\cite{ThomasQDMGG19}, point cloud ~\cite{WuFBDA19} and depth images~\cite{ParkJHLK20}.
In contrast to existing deformable kernels, the proposed 2D deformable kernel is manipulated by a 3D spatial pattern generator. The latter maps between the 3D space and the 2D image plane, while the formers work only within a single space. 

%% file: method.tex
\section{Method}
\label{sec:method}

\subsection{Overview}

Given an RGB image of an object, our goal is to reconstruct the complete 3D shape of the object with high-quality geometric details. 
We use signed distance fields (SDF) to represent the 3D objects and approximate the SDFs with neural network.
Our network takes 3D points $p=(x,y,z)\in \mathbb{R}^3$ and an image $I$ as input and outputs the signed distance $s$ at each input location. 
With an SDF, the surface of an object can be extracted as the isosurface of $SDF(\cdot)=0$ through the Marching Cubes algorithm. 
In general, our network consists of a fully convolutional image encoder $m$ and a continuous implicit function $f$ represented as multi-layer perceptrons (MLPs), from which the SDF is generated as
\begin{equation} \label{eq1}
f(p,F_l(a),F_g)=s, s\in\mathbb{R},
\end{equation}
where $a=\pi(p)$ is the 2D projection for $p$, $F_l(a) = m(I(a))$ is the local feature at image location $a$, and $F_g$ represents the global image feature. Feature $F_l(a)$ integrates the multi-scale local image features from the feature maps of $m$, from which the local image features are localized by aligning the 3D points to the image pixels via camera $c$.

By integrating with a spatial pattern at each 3D point sampling, the feature $F_l(a)$ of the sampling is modified by the local image features of the pattern points. We devise a feature encoding kernel $h$ attaching to the image encoder $m$ to encode a new local image feature from the features extracted from the image feature map.
Then our model is reformulated as
\begin{equation} \label{eq2}
f(p,h(F_l(a),F_l(a_1),...,F_l(a_n)),F_g)=s,
\end{equation}
where pixels $a_1,...a_n$ are the 2D projections of the 3D points $p_1,...,p_n$ belonging to the spatial pattern of the point sampling $p$. The encoding kernel $h$ is an MLP network that fuses the local image features. $n$ is the number of the pattern points.
Points $p_1,...,p_n$ are generated by a spatial pattern generator which is addressed in the following subsection.

In general, our pipeline is designed to achieve better exploitation of contextual information from local image features extracted according to the predicted 3D spatial patterns, resulting in geometry-sensitive image feature descriptions for 3D point samplings, ultimately improve the 3D reconstruction from single-view images.
A schematic illustration of the proposed model is given in Figure~\ref{fig:network}.

\begin{figure}
\centering
\includegraphics[width=1.0\linewidth]{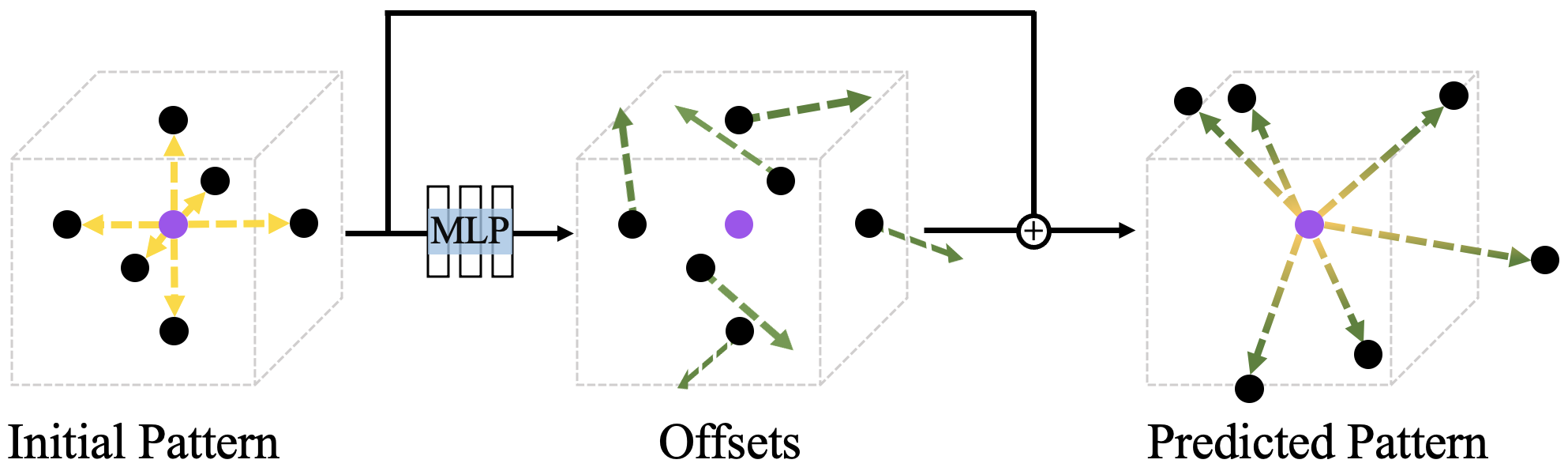}
\vspace{-8pt}
\caption{Illustration of spatial pattern generator. For an input point sampling, a pattern is initialized from $n$ nearby points around it, and the offsets of each surrounding points is predicted by an MLP network. The final pattern is created as the sum of initial points and the corresponding offsets.}
\label{fig:sp-generation}
\end{figure}

\subsection{Spatial Pattern Generator}\label{sec:spatial_pattern}

Our spatial pattern generator takes  as input a 3D point sampling $p$, and outputs $n$ 3D coordinates, i.e., $p_1,...,p_n$.
Like previous deformable convolution networks~\cite{dai2017deformable}, the position of a pattern point is computed as the sum of the initial location and a predicted offset. A schematic illustration of the spatial pattern generator is shown in Figure~\ref{fig:sp-generation}.

With proper initialization, the pattern can be learned efficiently and is highly effective for geometric reasoning.

\begin{figure}
\centering
\includegraphics[width=0.88\linewidth]{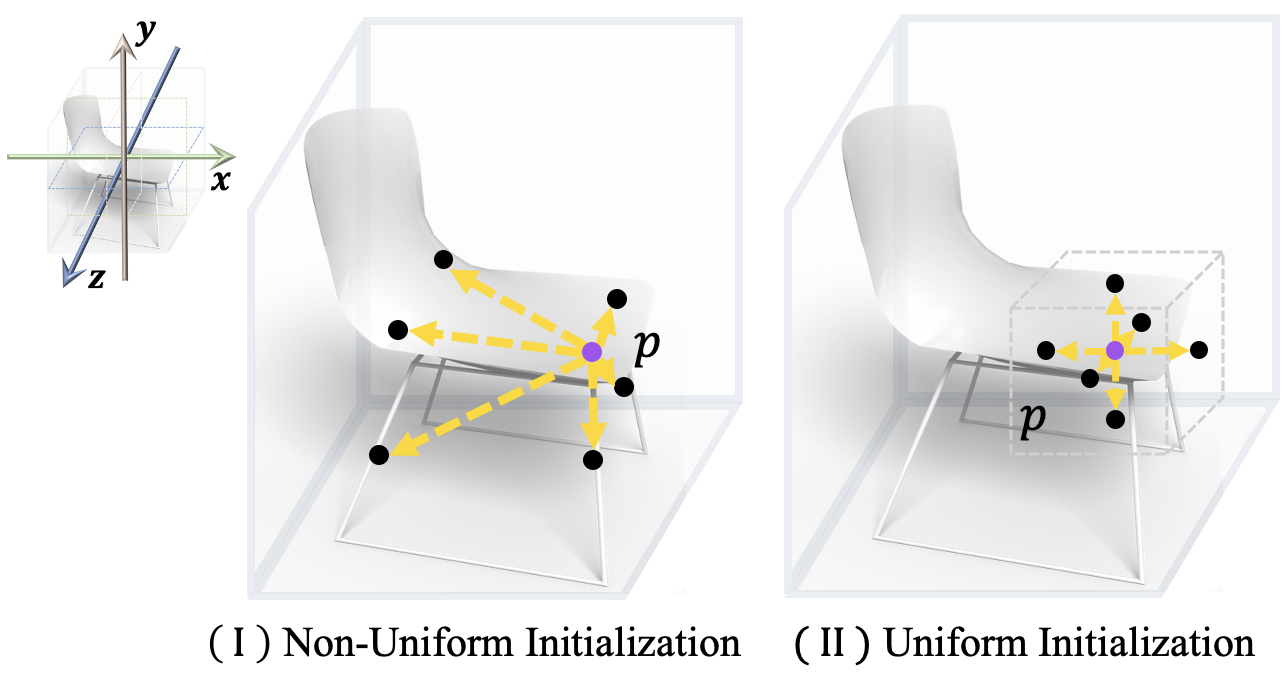}
\caption{Examples of spatial pattern initialization obtained by non-uniform sampling (I) and uniform sampling (II) strategies. In (I), a non-uniform pattern is formed by 3D points (in black) that are symmetry to input point $p$ along $x,y,z$ axis and $xy,yz,xz$ planes; and in (II), an uniform pattern is created by 3D points ling at centers of the side faces of a cube centered at point $p$.}
\label{fig:sp}
\end{figure}

\subsubsection{Initialization.} 
We consider two different sampling methods for spatial pattern initialization, i.e, uniform and non-uniform 3D point samplings. For simplicity, the input shapes are normalized to a unified cube centered at the origin.
For uniform pattern, we uniformly sample $n$ points from a cube centered at point $p=(x,y,z)$ with edge length of $l$. For example, we set $n=6$ and $l=0.2$, then the pattern points $p_i$ can be drawn from the combinations of $p_i=(x\pm 0.1,y\pm 0.1,z\pm 0.1)$, such that the pattern points lie at the center of the six side faces of the cube.

Unlike the uniform sampling method, the non-uniform sampling method does not have a commonly used strategy, except for random sampling. Randomly sampled points always do not have intuitive geometric meaning and are hardly appeared in any kernel point selection methods. As to capture non-local geometric relations, we design pattern points as the symmetry points of $p$ along global axes or planes that go through the origin of the 3D coordinate frame. Since the shape is normalized and centered at the origin, the symmetry points also lie within the 3D shape space.
Similarly, we let $n=6$, and we use $x, y, z$ axes and $xy, yz, xz$ planes to compute the symmetry points, then
the pattern points $p_i$ can be drawn from the combinations of $p_i=(\pm x,\pm y,\pm z)$.

The two different types of spatial pattern initialization are shown in Figure~\ref{fig:sp}.
After initialization, the pattern points, along with input sampling, are passed to an MLP network to generate the offset of each pattern point, and the final pattern is the sum of the initial pattern positions and the predicted offsets.

\paragraph{Loss Function.}
Given a collection of 3D shapes and the generated implicit fields from images $\mathcal{I}$, the loss is defined with $L_1$ distance:
\begin{equation} \label{eq4}
L_{SDF}=\sum_{I\in\mathcal{I}}\sum_{p} \omega  |f(p,F_l^I,F^I_g) -SDF^{I}(p) |,
\end{equation}
where $SDF^I$ denotes the ground truth SDF value corresponding to image $I$ and $f(\cdot)$ is the predicted field. $\omega$ is set to $\omega_1$, if $SDF^{I}(p)<\delta$, and $\omega_2$, otherwise.

%% file: results.tex
\section{Experiments} 
\label{sec:exp}

In this section, we show qualitative and quantitative results on single-view 3D reconstruction from our method, and comparisons with state-of-the-art methods. We also conduct a study on the variants of spatial patterns with quantitative results. To understand the effectiveness of the points in the spatial pattern, we provide an analysis of the learned spatial pattern with visualization and statistics.

\begin{table*}[h!]
\footnotesize
\tabcolsep=0.07cm
\begin{center}
\begin{tabular}{l|l|ccccccccccccc|c}
\hline
Metrics & Methods & plane & bench & cabinet & car& chair& display & lamp & speaker& rifle & sofa& table & phone & watercraft & mean \\ \hline
\multirow{3}{*}{\textbf{CD}$\downarrow$} & Ours$_{ uniform-6p}$ & 3.72 & 3.73 & 7.09 & 3.93 & 4.59 & 4.78 & 7.77 & 9.19 & 2.02 & 4.64 & 6.71 & 3.62 & 4.17 & 5.07 \\
& Ours$_{ non-uniform-6p}$ & 3.27 & 3.38 & 6.88 & 3.93 & 4.40 & 5.40 & 6.77 & 8.48 & 1.58 & 4.38 & 6.49 & 4.02 & 4.01 & \textbf{4.85} \\
& Ours$_{ non-uniform-3p}$ & 3.33 & 3.51 & 6.88 & 3.87 & 4.38 & 4.58 & 7.22 & 8.76 & 3.00 & 4.45 & 6.66 & 3.63 & 4.11 &  4.95 \\
\hline
\T\T 
\multirow{3}{*}{\textbf{EMD}$\downarrow$}  & Ours$_{ uniform-6p}$ & 2.07 & 2.02 & 2.60 & 2.38 & 2.19 & 2.11 & 2.86 & 2.85 & 1.55 & 2.16 & 2.41 & 1.78 & 2.01 & 2.23 \\
& Ours$_{ non-uniform-6p}$ & 1.91 & 1.90 & 2.58 & 2.36 & 2.17 & 2.08 & 2.66 & 2.75 & 1.52 & 2.11 & 2.36 & 1.77 & 1.99 & \textbf{2.17} \\
& Ours$_{ non-uniform-3p}$ & 1.96 & 1.94 & 2.58 & 2.35 & 2.16 & 2.07 & 2.81 & 2.81 & 1.58 & 2.13 & 2.39 & 1.78 & 2.00 & 2.20  \\
\hline
\T\T
\multirow{3}{*}{\textbf{IOU}$\uparrow$} & Ours$_{ uniform-6p}$ & 66.1  & 59.5  &  59.6 &   80.0 & 65.8  &  66.7 & 53.8  & 63.7 &  74.7 & 74.1 &  60.8 & 79.6  & 68.0  & 67.1 \\
&Ours$_{ non-uniform-6p}$ & 68.2  & 63.1  & 61.4   &  80.7   & 66.8  &  67.9 & 55.9   &  65.0 &  75.0  & 75.2 &  62.6 & 81.0  & 68.9  & \textbf{68.6}  \\
&Ours$_{ non-uniform-3p}$ & 67.4  & 62.0  &  60.5 &   80.5 & 66.8  &  67.5 & 54.1  & 64.2 &  73.6 & 75.1 &  61.8 & 80.2  & 68.7  & 67.9  \\
\hline
\end{tabular}
\caption{Quantitative results of the variants of our method using different configurations of spatial pattern. 
Metrics include CD (multiply by 1000, the smaller the better), EMD (multiply by 100, the smaller the better), and IoU (\%, the larger the better). CD and EMD are computed on 2048 points.}
\label{table:qv}
\end{center}

\end{table*}

\begin{table*}[h!]
\footnotesize
\tabcolsep=0.095cm
\begin{center}
\begin{tabular}{l|l|ccccccccccccc|c}
\hline
Metrics & Methods & plane & bench & cabinet & car& chair& display & lamp & speaker& rifle & sofa& table & phone & watercraft & mean \\ \hline
\multirow{6}{*}{\textbf{CD}$\downarrow$} 
&Pixel2Mesh & 6.10  & 6.20  & 12.11  & 13.45  & 11.13  & \textbf{6.39}  & 31.41  & 14.52  & 4.51  & \textbf{6.54}  & 15.61  & 6.04  & 12.66  & 11.28 \\
&OccNet & 7.70  & 6.43  & 9.36  & 5.26  & 7.67  & 7.54  & 26.46  & 17.30  & 4.86  & 6.72  & 10.57  & 7.17  & 9.09  & 9.70 \\
&DISN & 9.96 & 8.98 & 10.19 & 5.39 & 7.71 & 10.23 & 25.76 & 17.90 & 5.58 & 9.16 & 13.59 & 6.40 & 11.91 & 10.98 \\
&Ladybird & 5.85 & 6.12 & 9.10 & 5.13 & 7.08 & 8.23 & 21.46 & 14.75 & 5.53 & 6.78 & 9.97 & \textbf{5.06} & 6.71 & 8.60 \\
&Ours$_{cam}$ & \textbf{5.40} & \textbf{5.59} & 8.43 & \textbf{5.01} & \textbf{6.17} & 8.54 & 14.96 & \textbf{14.07} & 3.82 & 6.70 & \textbf{8.97} & 5.42 & \textbf{6.19} & \textbf{7.64}\\
&Ours & 3.27 & 3.38 & 6.88 & 3.93 & 4.40 & 5.40 & 6.77 & 8.48 & 1.58 & 4.38 & 6.49 & 4.02 & 4.01 & 4.85 \\
\hline
\T\T 
\multirow{6}{*}{\textbf{EMD}$\downarrow$} 
&Pixel2Mesh & 2.98  & 2.58  & 3.44  & 3.43  & 3.52  & 2.92  & 5.15  & 3.56  & 3.04  & 2.70  & 3.52  & 2.66  & 3.94  & 3.34 \\
&OccNet & 2.75  & 2.43  & 3.05  & 2.56  & 2.70  & 2.58  & 3.96  & 3.46  & 2.27  & \textbf{2.35}  & 2.83  & 2.27  & 2.57  & 2.75 \\
&DISN & 2.67  & 2.48  & 3.04  & 2.67  & 2.67  & 2.73 &  4.38 &  3.47 &  2.30  & 2.62  & 3.11  & 2.06 &  2.77  & 2.84 \\
&Ladybird & 2.48  & 2.29  & 3.03  & 2.65 &  2.60  & 2.61 & 4.20  & 3.32  & 2.22  & 2.42 &  2.82  & 2.06 &  2.46  & 2.71 \\
&Ours$_{cam}$ &\textbf{2.35} & \textbf{2.15} & \textbf{2.90} & 2.66 & \textbf{2.49} & \textbf{2.49} & \textbf{3.59} & \textbf{3.20} & \textbf{2.04} & 2.40 & \textbf{2.70} & \textbf{2.05} & \textbf{2.40}  & \textbf{2.57} \\
&Ours & 1.91 & 1.90 & 2.58 & 2.36 & 2.17 & 2.08 & 2.66 & 2.75 & 1.52 & 2.11 & 2.36 & 1.77 & 1.99 & 2.17 \\
\hline
\T\T
\multirow{6}{*}{\textbf{IOU}$\uparrow$} 
&Pixel2Mesh & 51.5  & 40.7  & 43.4  & 50.1  & 40.2  & 55.9  & 29.1  & 52.3  & 50.9  & 60.0  & 31.2  & 69.4 &  40.1  & 47.3\\
&OccNet & 54.7  & 45.2  & \textbf{73.2}  & 73.1  & 50.2  & 47.9  & 37.0  & \textbf{65.3}  & 45.8  & 67.1  & \textbf{50.6} &  70.9  & 52.1  & 56.4 \\
&DISN & 57.5  & 52.9  & 52.3  & 74.3  & 54.3 &  56.4  & 34.7 &  54.9  & 59.2  & 65.9  & 47.9 &  72.9 &  55.9  & 57.0 \\
&Ladybird & 60.0  & 53.4 &  50.8  & 74.5  & 55.3  & 57.8  & 36.2  & 55.6 &  61.0  & 68.5  & 48.6  & 73.6 &  \textbf{61.3}  & 58.2 \\
&Ours$_{cam}$ & \textbf{60.6} & \textbf{54.4} & 52.9 & 74.7 & \textbf{56.0} & \textbf{59.2} & \textbf{38.3} & 56.1 & \textbf{62.9} & \textbf{68.8} & 49.3 & \textbf{74.7} & 60.6 & \textbf{59.1} \\
&Ours & 68.2  & 63.1  & 61.4 &  80.7   & 66.8  &  67.9 & 55.9   &  65.0 &  75.0  & 75.2 &  62.6 & 81.0  & 68.9  & 68.6  \\
\hline
\end{tabular}
\caption{
Quantitative results on the ShapeNet Core dataset for various methods.}
\label{table:qr}
\end{center}
\vspace{-6pt}

\end{table*}

\paragraph{Implementation Details.}
We use DISN~\cite{DISN} as our backbone network, which consists of a VGG-style fully convolutional neural network as the image encoder and multi-layer perceptrons (MLPs) to represent the implicit function. 
Our spatial pattern generator and local feature aggregation kernel are based on MLPs.
The parameters of our loss function in Equation~\ref{eq4} are set to $\omega_1=4,\omega_2=1,$ and $\delta = 0.01$. 

\paragraph{Dataset and Training Details.}
We use the ShapeNet Core dataset~\cite{ShapeNet2015} and Pix3D dataset~\cite{pix3d} for evaluation. 
The proposed model is trained across all categories. 
In testing, for the ShapeNet dataset, the camera parameters are estimated from the input images, and we use the trained camera model from DISN~\cite{DISN} for fair comparisons. For the Pix3D dataset, ground truth camera parameters and image masks are used.

\paragraph{Evaluation Metrics.}
The quantitative results are obtained by computing the similarity between generated surfaces and ground truth surfaces. We use the standard metrics including Chamfer Distance (CD), Earth Mover’s Distance (EMD), and Intersection over Union (IoU). 

\T\T
Please refer to the supplemental materials for more details on data processing, network architecture, training procedures, and spatial pattern computation.

\subsection{Quantitative and Qualitative Evaluations}
\label{subsec:qte}

\subsubsection{Impact of Spatial Pattern Configuration.}
To figure out the influence of different spatial patterns, we designed several variants of the pattern. Specifically, two factors are considered, including the initialization and the capacity, i.e., the pattern point sampling strategy and the number of the points in a pattern. As described before, we considers non-uniform and uniform sampling methods for pattern initialization and set the number of points to three to six for changing the capacity. The methods derived from the combinations of the aforementioned two factors are denoted as

\begin{itemize}[itemsep=2pt,topsep=0pt,parsep=2pt]
\item Ours$_{uniform-6p}$, in which six points are uniformly sampled on a cube $l$ centered at point sampling.
\item Ours$_{non-uniform-6p}$, in which six points are non-uniformly sampled at the symmetry locations in the shape space along $xy,yz$ and $xz$ planes and $x,y$ and $z$ axes.
\item Ours$_{non-uniform-3p}$, in which three points are non-uniformly sampled at the symmetry locations in the shape space along $xy,yz$ and $xz$ planes.
\end{itemize}


In Table~\ref{table:qv}, we report the numerical results of the methods using ground truth camera pose. In general, Ours$_{non-uniform-6p}$ achieves best performance. By reducing the capacity to the number of three points, the performance decreases, as shown by Ours$_{non-uniform-3p}$. This indicates that some critical points in Ours$_{non-uniform-6p}$ that have high responses to the query point do not appear in Ours$_{non-uniform-3p}$. 
Notably, the sampling strategy is more important. Both Ours$_{non-uniform-6p}$ and Ours$_{non-uniform-3p}$ outperforms Ours$_{uniform-6p}$ with large margins. Thus, initialization with non-uniform sampling makes the learning of effective spatial patterns easier. This implies that optimizing the pattern position in the continuous 3D space is challenging, and with proper initialization, the spatial pattern can be learned more efficiently. To better understand the learned spatial pattern and which pattern points are preferred by the network, we provide analysis with visualization and statistics in the next section. Before that, we evaluate the performance of our method by comparing it with several state-of-the-art methods. Specifically, we use Ours$_{non-uniform-6p}$ as our final method for comparison.

\subsubsection{Comparison with Various Methods.}
We compare our method with the state-of-the-art methods on the single-image 3D reconstruction task. 
All the methods, including 
OccNet~\cite{OccNet}, 
DISN~\cite{DISN}, Ladybird~\cite{xu2020ladybird}, are trained across all 13 categories. 
The method of Ours uses ground truth cameras while Ours$_{cam}$ denotes the version of Ours using estimated camera poses.

A quantitative evaluation on the ShapeNet dataset is reported in Table~\ref{table:qr} in terms of CD, EMD, and IOU. CD and EMD are evaluated on the sampling points from the generated triangulated mesh. IOU is computed on the solid voxelization of the mesh. In general, our method outperforms other methods. Particularly, among the methods including DISN, Ladybird, and Ours, which share a similar backbone network, Ours achieves much better performance. 

\begin{figure*}[h!]
\begin{center}
{\includegraphics[width=0.93\linewidth 
]
{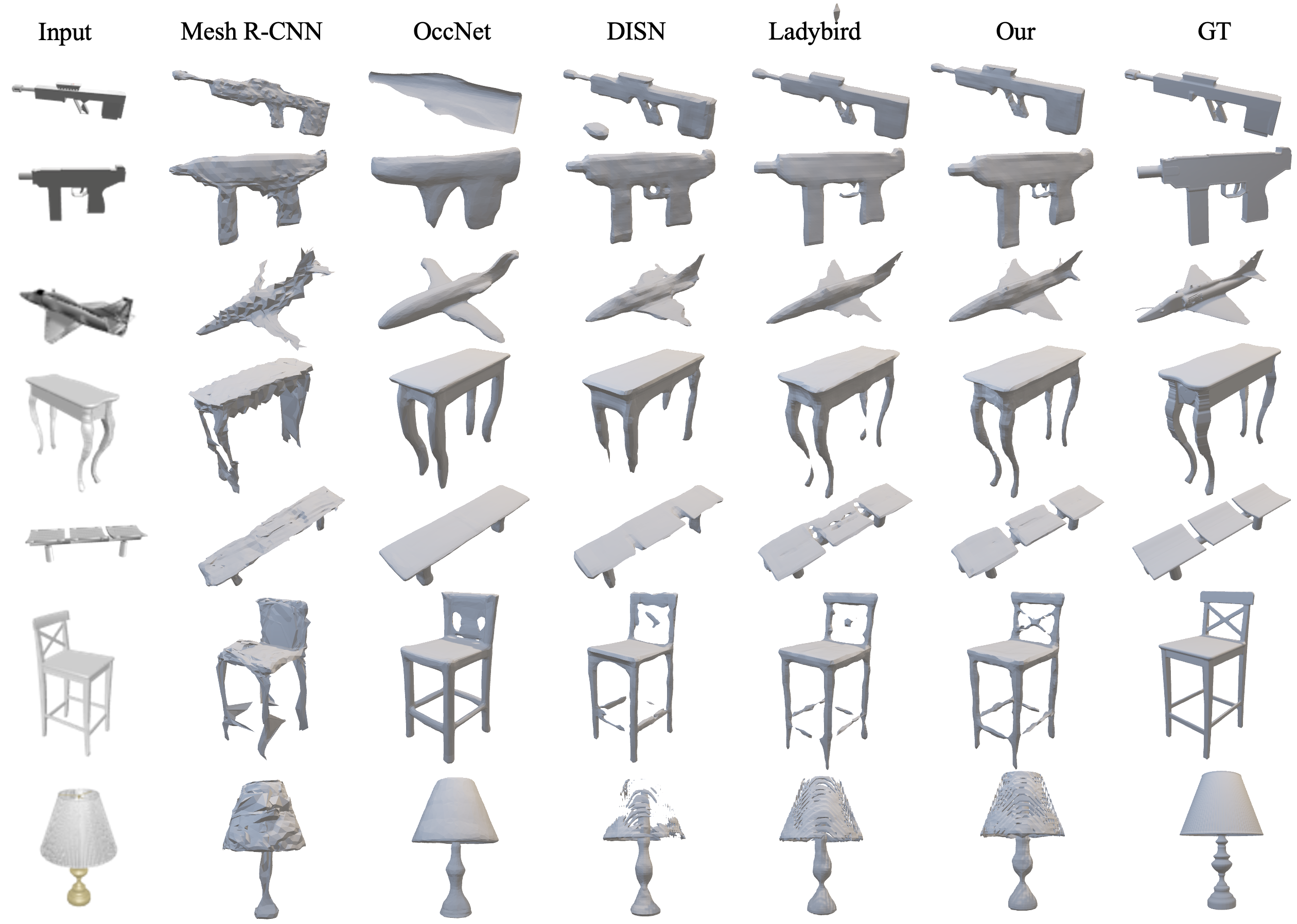} 
}
\end{center}

\caption{Qualitative comparison results for various methods.
}
\label{fig:results}
\vspace{-6pt}

\end{figure*}

In Figure~\ref{fig:results}, we show qualitative results generated by Mesh R-CNN~\cite{gkioxari2019mesh}, OccNet~\cite{OccNet}, DISN~\cite{DISN} and Ladybird~\cite{xu2020ladybird}. We use the pre-trained models from the Mesh R-CNN, OccNet, and DISN. For Ladybird, we re-implement their network and carry out training according to the specifications in their paper.
All the methods are able to capture the general structure of the shapes, shapes generated from DISN, Ladybird and Ours are more aligned with the ground truth shapes.  
Specifically, our method is visually better at the non-visible regions and fine geometric variations.

\begin{table}[h!]
\footnotesize
\tabcolsep=0.12cm
\begin{center}
\begin{tabular}{l|c|c|c|c|c|c}
         & \multicolumn{2}{c|}{\textbf{CD}(x1000)$\downarrow$} & \multicolumn{2}{c|}{\textbf{EMD}(x100)$\downarrow$} & \multicolumn{2}{c}{\textbf{IOU}(\%)$\uparrow$} \\\hline
         & Ladybird     & Ours    & Ladybird     & Ours     & Ladybird     & Ours \\\hline
bed      & 9.84         & \textbf{8.76}    & 2.80         & \textbf{2.70}     &  70.7        & \textbf{73.2}   \\
bookcase & \textbf{10.94}        & 14.70   & \textbf{2.91}         & 3.32     &  \textbf{44.3}        & 41.8    \\
chair    & 14.05        & \textbf{9.81}    & 2.82         & \textbf{2.72}     &  \textbf{57.3}        & \textbf{57.3}     \\
desk     & 18.87        & \textbf{15.38}   & 3.18         & \textbf{2.91}     &  51.2        & \textbf{60.7}     \\
misc     & 36.77        & \textbf{30.94}   & 4.45         & \textbf{4.00}     &  29.8        & \textbf{44.0}  \\
sofa     & 4.56         & \textbf{3.77}   & 2.02         & \textbf{1.92}     &  86.7        & \textbf{87.6}     \\
table    & 21.66        & \textbf{14.04}   & 2.96         & \textbf{2.78}     &  56.9        & \textbf{58.8}      \\
tool     & \textbf{7.78}         & 16.24   & 3.70         & \textbf{3.57}     &  \textbf{41.3}        & 38.2     \\
wardrobe & \textbf{4.80}         & 5.60    & \textbf{1.92}         & 2.01     &  \textbf{87.5}        & \textbf{87.5}     \\\hline
mean     & 14.36        & \textbf{13.25}   & 2.97         & \textbf{2.88}     &  58.4        & \textbf{61.0}    
\end{tabular}
\caption{Quantitative results on Pix3D dataset.}
\label{table:pix3d}
\end{center}
\end{table}

The quantitative evaluation of the Pix3D dataset is provided in Table~\ref{table:pix3d}.
Both Ours and Ladybird are trained and evaluated on the same train/test split, during which ground truth camera poses and masks are used. Specifically, $80\%$ of the images are randomly sampled from the dataset for training while the rest images are used for testing. In general, our method outperforms Ladybird on the used metrics.

\begin{figure}[h!]
\centering
\includegraphics[width=0.95\linewidth]{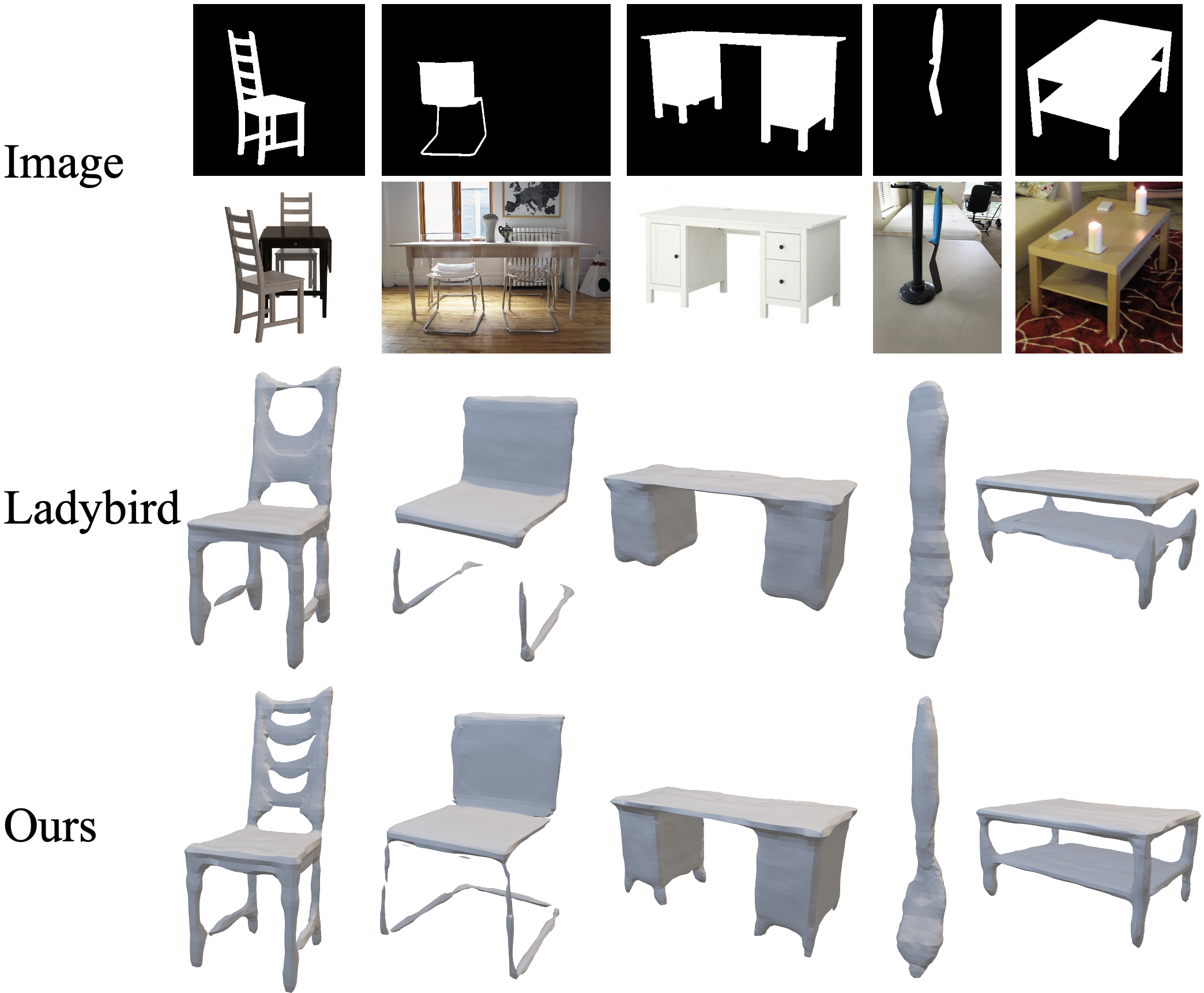}
\caption{Qualitative Results on the Pix3D dataset. Ground truth image masks and camera parameters are used.}
\label{fig:pix3d}
\end{figure}
In addition to the quantitative results, we also show the reconstructed shapes in Figure~\ref{fig:pix3d}. Compared to the synthetic images from the ShapeNet dataset, the real images are more diverse in terms of camera views, object sizes, and appearances. Our reconstruct shapes are visually more plausible compared to Ladybird.

\subsection{Analysis of Learned Spatial Patterns}
\label{subsec:ana}

\begin{figure}
\centering
\includegraphics[width=1.0\linewidth]{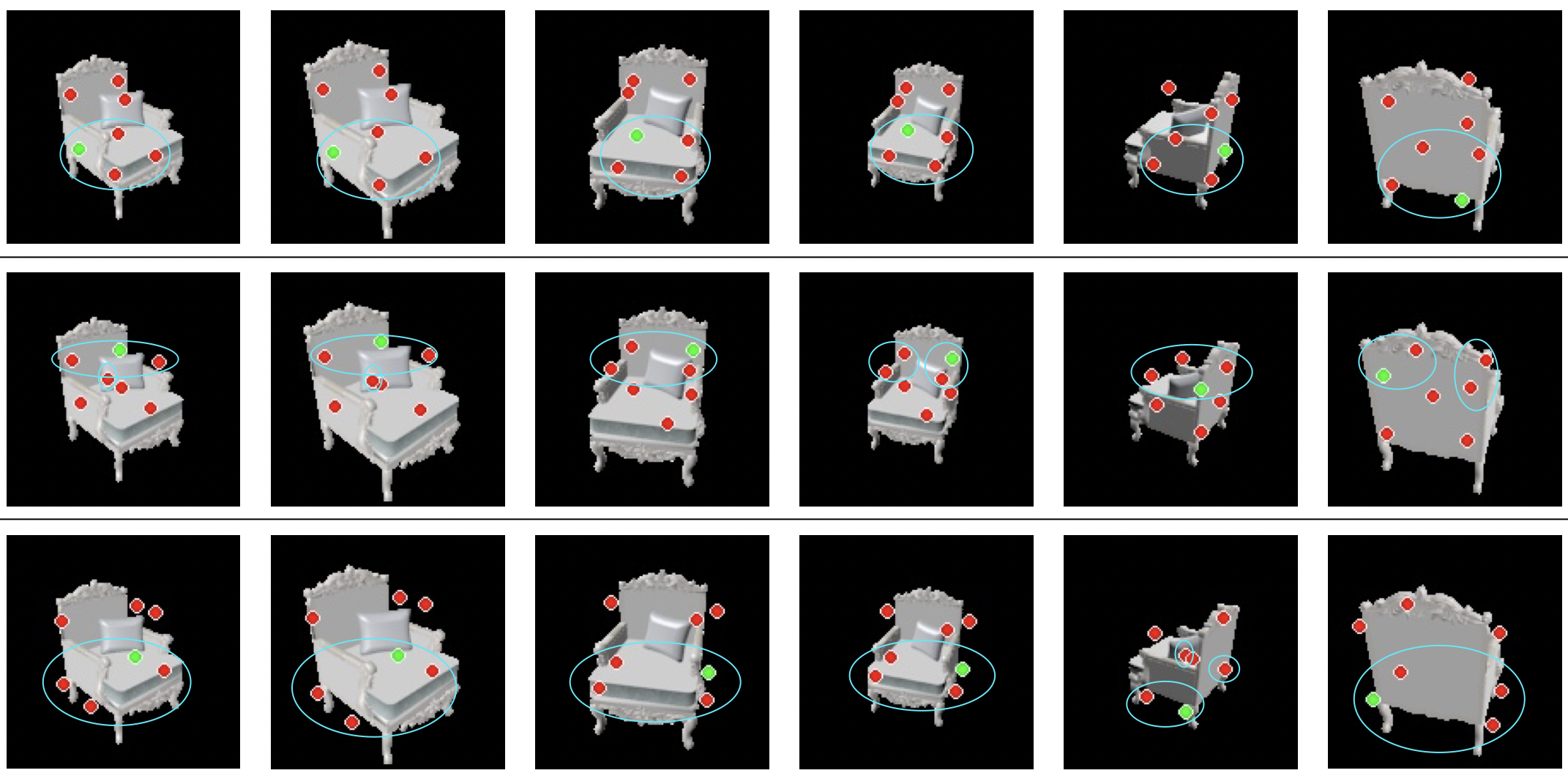}
\vspace{-10pt}
\caption{Visualization of learned spatial patterns in image plane. 
}
\label{fig:2d-offset}
\end{figure}

\begin{figure}
\centering
\includegraphics[width=1.0\linewidth]{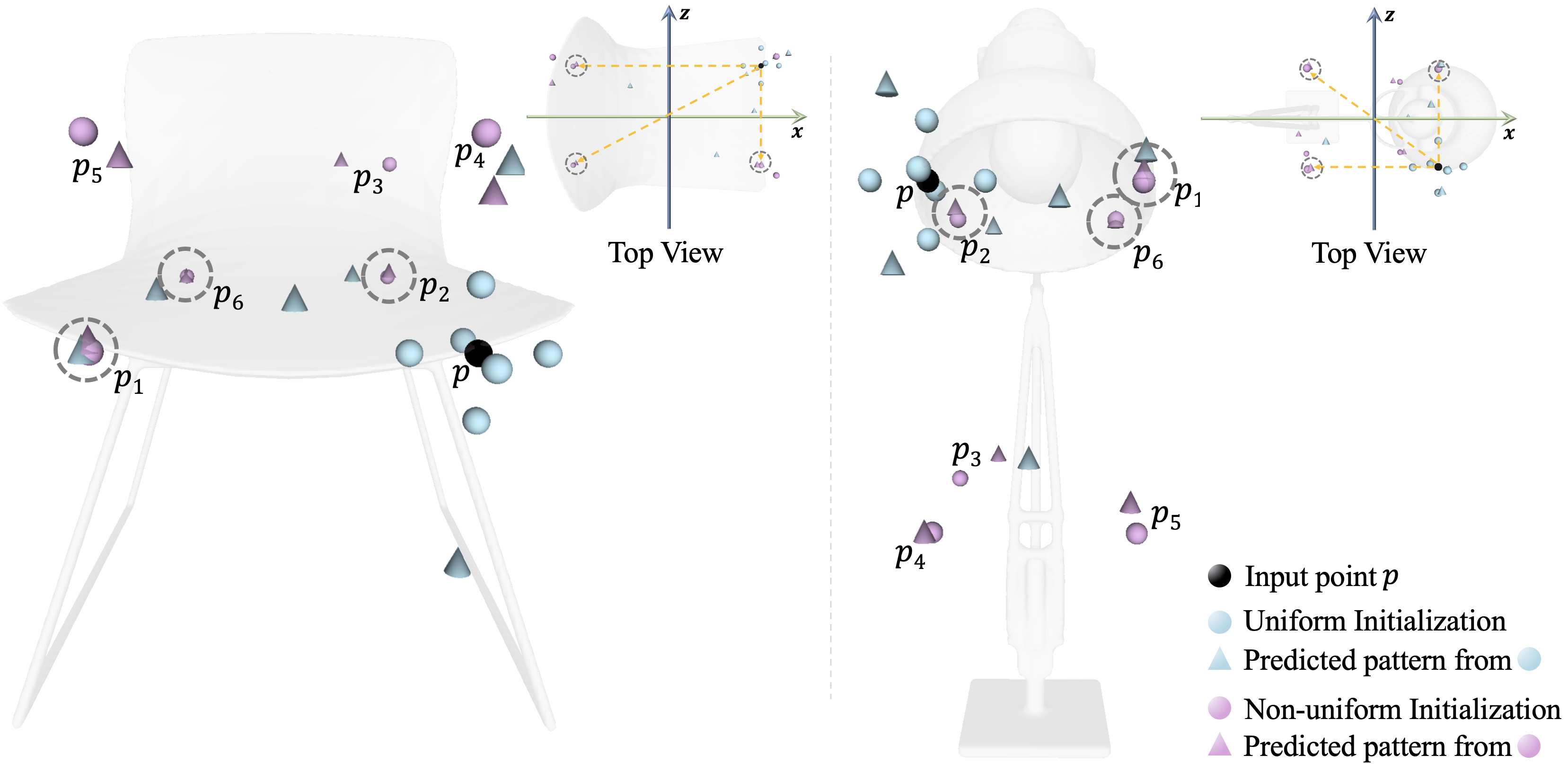}
\vspace{-10pt}
\caption{Visualization of spatial pattern points with different shapes and colors. 
From the examples in (I) and (II), the learned pattern points (i.e., pink cones) from the non-uniform initialization (i.e., pink balls) are relative stationary, while points (i.e., blue cones) learned from uniform initialization (i.e., blue balls) have much larger deviations from their original positions. Some stationary points $p_1,p_2$ and $p_6$ are highlighted in dash circles.}
\label{fig:sp-vis}
\end{figure}

\begin{figure}
\centering
\includegraphics[width=1.0\linewidth]{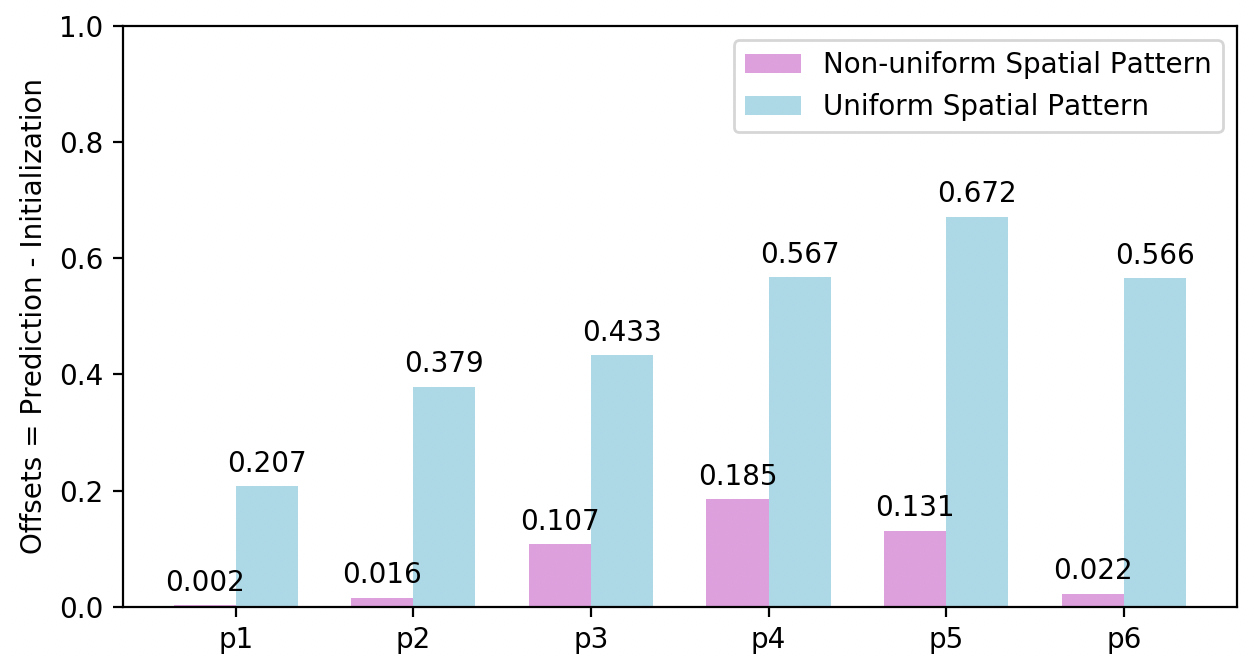}
\vspace{-12pt}
\caption{Statistics on the offsets of spatial pattern points. The offset of individual pattern point is computed as the mean distance between the initial and predicted position. Among all points, $p_1,p_2$ and $p_6$ have smallest learned offsets from the non-uniform initialization (i.e., pink bars), while for uniform initialization (i.e., blue bars), all the predicted points have much larger deviations from their original locations.}
\label{fig:sp-stat}
\vspace{-6pt}
\end{figure}

We have demonstrated the effectiveness of the proposed spatial pattern via achieving better performance than other alternatives, and the experiments on different variants of the spatial pattern show the influence of initialization and capacity. To better understand the importance of individual pattern points, we visualize several learned patterns in Figure~\ref{fig:2d-offset}\&\ref{fig:sp-vis} and calculate the mean offsets of the predicted pattern points visualized in Figure~\ref{fig:sp-stat}.

In Figure~\ref{fig:2d-offset}, we show learned spatial patterns in the 2D image plane. In each row, a spatial pattern is shown in six different images with different views. This implies an explicit constraint of view consistency on image encoding. 

Pattern points (colored in red) that have intuitive geometric relationships (e.g., symmetric and co-planar) with the query points (colored in green) are highlighted by cyan circles in Figure~\ref{fig:2d-offset}.
Figure~\ref{fig:sp-vis} provides a better visualization in 3D frame, from which we can see that some learned pattern points from the non-uniform initialization are almost stationary, e.g., points $p_1,p_2$ and $p_6$ that are highlighted by dash circles. Also, as shown in Figure~\ref{fig:sp-stat}, the mean offsets of points $p_1,p_2$ and $p_6$ are close to zero. 
To figure out the importance of these stationary pattern points, we train the network using the points $p_1,p_2$, and $p_6$ as a spatial pattern and keep their positions fixed during training. As shown in Table~\ref{table:select-sp}, the performance of the selected rigid pattern is better than Ours$_{non-uniform-3p}$ and Ours$_{uniform-6p}$ and slightly lower than Ours$_{non-uniform-6p}$. This reveals that the pattern points discovered by the network are useful, which finally lead to a better reconstruction of the underlying geometry.

\begin{table*}
\footnotesize
\tabcolsep=0.145cm
\begin{center}
\begin{tabular}{l|ccccccccccccc|c}
\hline
 & plane & bench & cabinet & car& chair& display & lamp & speaker& rifle & sofa& table & phone & watercraft & mean \\ \hline
\textbf{CD}(x1000) & 3.38 & 3.44 & 7.06 & 3.87 & 4.50 & 4.57 & 7.30 & 8.98 & 1.66 & 4.53 & 6.61  & 3.45 & 4.17 &  4.89 \\
\hline
\textbf{EMD}(x100) & 1.97 & 1.92 & 2.58 & 2.37 & 2.16 & 2.07 & 2.77 & 2.82 & 1.52 & 2.12 & 2.35  & 1.80 & 2.01 & 2.19  \\
\hline
\textbf{IOU}(\%) & 67.4  & 62.8  &  60.5 &   80.5 & 66.6  &  67.4 & 54.9  & 64.5 &  74.9 & 75.0 & 62.5  & 80.1  & 68.5  & 68.1  \\
\hline
\end{tabular}
\caption{Quantitative results of a rigid spatial pattern formed by three pattern points selected from the stationary points of the learned spatial pattern.}
\label{table:select-sp}
\end{center}
\end{table*}

Even though consuming more time and memory, utilizing the auxiliary contextual information brought by other points $p_3,p_4$, and $p_5$ only achieve slight improvement on the performance. 
The analysis shows that 
naive selection of more neighboring points is not as effective as the strategy that considers the underlying geometric relationships. Although there is no explicit constraint to guarantee the geometric relations exactly, statistically we found that the network tends to shift the pattern points towards the locations that have geometric relations with the query point, as shown in~\ref{fig:sp-vis}\&\ref{fig:sp-stat}.
It further proves that encoding geometric relationships with the 2D kernel derived by the proposed spatial pattern is effective for the single-image 3D reconstruction task.

%% file: conclusion.tex
\section{Conclusion}

In this paper, we propose a new feature encoding scheme working on deep implicit field regression models for the task of 3D shape reconstruction from single images. We present spatial pattern, from which a kernel operating in image space is derived to better encode local image features of 3D point samplings.
Using spatial pattern enables the 2D kernel point selection explicitly to consider the underlying 3D geometry relations, which are essential in 3D reconstruction task, while traditional 2D kernels mainly consider the appearance information.
To better understand the spatial pattern, we study several variants of spatial pattern designs in regard to the pattern capacity and the way of initialization, and we analyze the importance of individual pattern points.
Results on large synthetic and real datasets show the superiority of the proposed method on widely used metrics.

A key limitation is that the model is sensitive to camera parameters. As shown in Table~\ref{table:qr}, when camera parameters are exactly correct, the performance is significantly improved. One possible direction to investigate is to incorporate the camera estimation process in the loop of 3D reconstruction pipeline, such as jointly optimize the camera pose and the implicit field within a framework with multiple objectives.
Another interesting direction is to learn geometric relations with explicit geometric constraints. Restricting the optimization to an optimized subspace could potentially promote performance and interpretation of learned patterns.

%% file: appendix.tex
\newpage
\clearpage

\section{Appendix}

\subsection{Data Processing}

\subsubsection{Datasets.}
The ShapeNet Core dataset~\cite{ShapeNet2015} includes 13 object categories, and for each object, 24 views are rendered with resolution of 137$\times$137 as in~\cite{choy20163d}. 
Pix3D Dataset~\cite{pix3d} contains 9 object categories with real-world images and the exact mask images. The number of views and the image resolution varies from different shapes.We process all the shapes and images in the same format for the two datasets. Specifically, all shapes are normalized to the range of [-1,1] and all images are scaled to the resolution of 137$\times$137.

\subsubsection{3D Point Sampling.}
For each shape, 2048 points are sampled for training. We firstly normalize the shapes to a unified cube with their centers of mass at the origin. Then we uniformly sample $256^3$ grid points from the cube and compute the sign distance field (SDF) values for all the grid samples. Following the sampling process of Ladybird~\cite{xu2020ladybird}, the $256^3$ points are downsampled with two stages. 
In the first stage, 32,768 points are randomly sampled from the four SDF ranges [-0.10,-0.03], [-0.03,0.00], [0.00,0.03], and [0.03,0.10], with the same probabilities.
In the second stage, 2048 points are uniformly sampled from the 32,768 points using the farthest points sampling strategy.

In testing, $65^3$ grid points are sampled are fed to the network, and output the SDF values. The object mesh is extracted as the zero iso-surface of the generated SDF using the Marching Cube algorithm.

\subsubsection{3D-to-2D Camera Projection.}
The pixel coordinate $a$ of a 3D point sampling $p$ is computed as two stages. Firstly, the point is converted from the world coordinate system to the local camera coordinate system $c$ based on the rigid transformation matrix $A^c$, such that $p^c = A^cp$. Then in the camera space, point $p^c=(x^c,y^c,z^c)$ is projected to the 2D canvas via perspective transformation, i.e., $\pi(p^c) = (\frac{x^c}{z^c},\frac{y^c}{z^c})$. The projected pixel whose coordinate lies out of an image will reset to 0 or 136 (the input image resolution is fixed as 137$\times$137 in our experiment). 

\begin{figure}[h!]
\begin{center}
{\includegraphics[width=0.99\linewidth]
{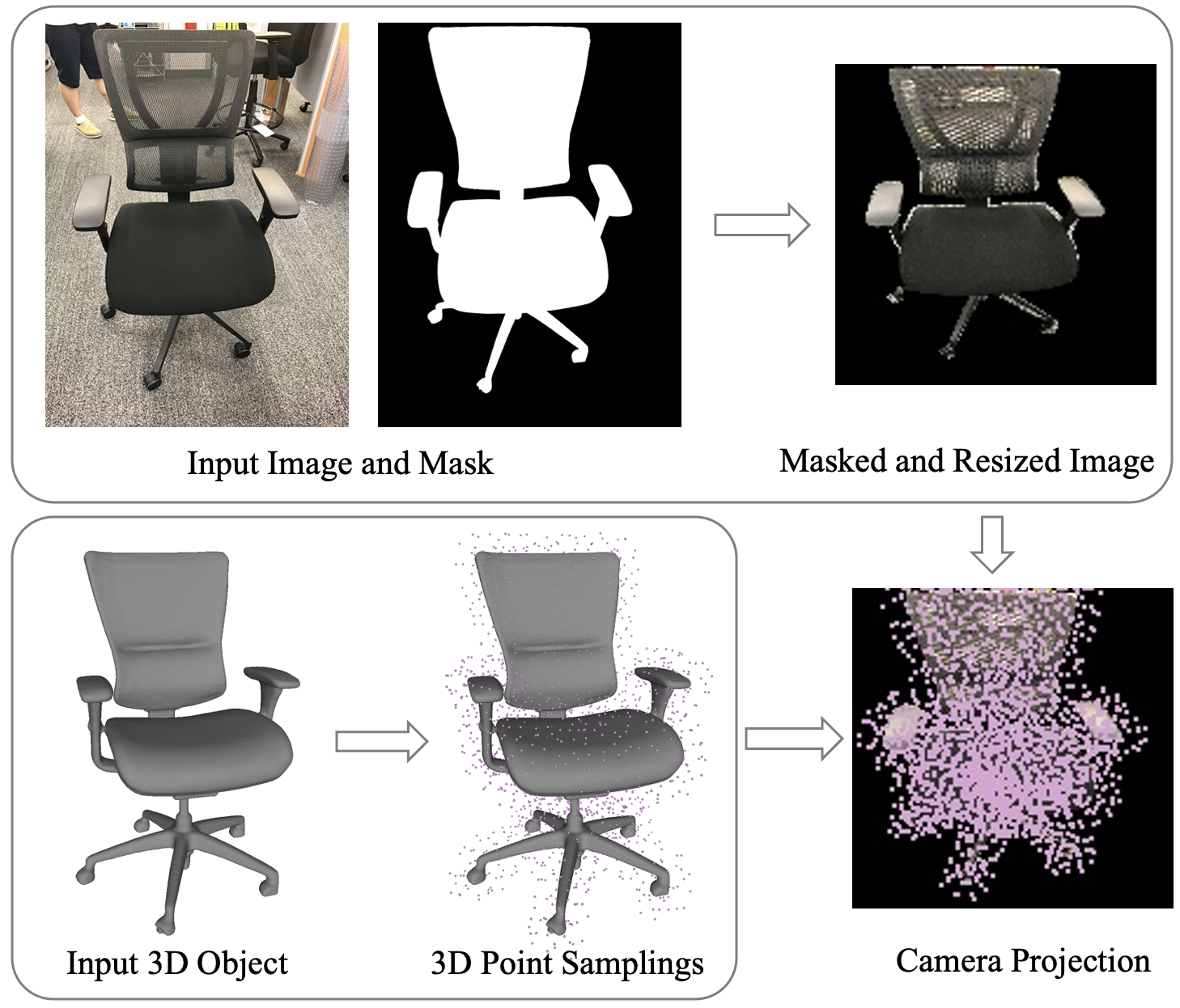} 
}
\end{center}
\vspace{-6pt}
\caption{
The data processing workflow, including 3D point sampling, image preparation and 3D-to-2D camera projection.
}
\label{fig:process}
\vspace{-6pt}
\end{figure}

\T\T\T
We show the data processing workflow with an example from the Pix3D dataset in Figure~\ref{fig:process}. For the ShapeNet dataset, the images are already satisfied.

\subsection{Spatial Pattern Computation}
\label{subsec:spg}

For a 3D point sampling $p=(x,y,z)$, the pattern points of $p$ are drawn from $p_i=(x\pm l/2,y\pm l/2,z\pm l/2)$ when using the uniform initialization, and are computed by $p_i=(\pm x,\pm y,\pm z)$ if adopting the non-uniform initialization, where $i$ is the point index, such that $i=1,...,n$, and $n$ is the number of pattern points. Specifically, the spatial patterns for the variants of our methods are computed as follows
\begin{itemize}
\item For Ours$_{uniform-6p}$, the pattern configuration parameters are set to $n=6$ and $l=0.2$, and the pattern points are $p_1= (x,y,z+0.1)$, $p_2= (x+0.1,y,z)$, $p_3= (x,y+0.1,z)$, $p_4= (x,y,z-0.1)$, $p_5= (x-0.1,y,z)$, and $p_6= (x,y-0.1,z)$.
\item For Ours$_{non-uniform-6p}$, the pattern configuration parameters are set to $n=6$, and the pattern points are $p_1= (x,y,-z)$, $p_2= (-x,y,z)$, $p_3= (x,-y,z)$, $p_4= (-x,-y,z)$, $p_5= (x,-y,-z)$, and $p_6= (-x,y,-z)$.
\item For Ours$_{non-uniform-3p}$, the pattern configuration parameters are set to $n=3$, and the pattern points are $p_1= (x,y,-z)$, $p_2= (-x,y,z)$, and $p_3= (x,-y,z)$.
\end{itemize}

As introduced in the analysis section, the rigid spatial pattern formed by the selected stationary points uses points $p_1,p_2$ and $p_6$ from the pattern points of Ours$_{non-uniform-6p}$.

\subsection{Network and Training Details}

\subsubsection{Traning Policy.}
We implement our method based on the framework of Pytorch. For training on the ShapeNet dataset, we use the Adam optimizer with a learning rate 1e-4, a decay rate of 0.9, and a decay step size of 5 epochs. The network was trained for 30 epochs with batch size 20.
For training on the Pix3D dataset, we use the Adam optimizer with a constant learning rate 1e-4, and smaller batch size 5.
For the ShapeNet dataset, at each epoch, we randomly sample a subset of images from each category. Specifically, a maximum number of 36000 images are sampled for each category. The total number of images in an epoch is 411,384 resulting in 20,570 iterations.
Our model is trained across all categories.

\subsubsection{Network Architecture.}
We introduce the details of the image encoder $m$, spatial pattern generator $g$, feature aggregation module $h$, and the SDF regression network $f$ in our paper.

We use the convolution network of VGG-16 as our image encoder, which generates multi-resolution feature maps. Similar to DISN~\cite{DISN}, we reshape the feature maps to the original image size with bilinear interpolation and collect the local image features of a pixel from all scales of feature maps. Specifically, the local feature contains six sub-features from the six feature maps, with the dimension of \{64, 128, 256, 512, 512, 512\} respectively.

\begin{table}[h!]
\footnotesize
\centering
\label{table:supp-network-sdf}
\begin{tabular}{c|c}
\hline
\multicolumn{2}{c}{\T  \bf Spatial Pattern Generator $g$} \\ \hline 
\multicolumn{2}{c}{Input $1+n$ $3$-dim points}\\ 
\multicolumn{2}{c}{$\rightarrow$ output $n$ $3$-dim points}\\ \hline
Operation & Output Shape \\ \hline
Conv1D+ReLU & (64,)\\\hline
Conv1D+ReLU & (256,) \\\hline
Conv1D+ReLU & (512,) \\\hline
Concat(\{$1+n$ features\}) & ($(1+n)\times 512$,)\\\hline
Conv1D+ReLU & (512,) \\\hline
Conv1D+ReLU & (256,) \\\hline
Conv1D+Tanh & ($n\times 3$,) \\\hline
\end{tabular}
\vspace{-6pt}
\caption{Spatial Pattern Generator.}
\end{table}

The spatial pattern generator receives $n+1$ local image features from a point sampling and an initial spatial pattern belonging to it. All the points are firstly promoted from $R^3$ to the dimension of $512$ using a multi-layer perceptron (MLP). Then all the point features are concatenated and passed to another MLP to output a vector with the dimension of $n\times 3$ resulting in $n$ 3D offset positions. The final pattern is the sum of the initial pattern and the predicted offsets.

\begin{table}[h!]
\footnotesize
\centering
\label{table:supp-network-feat}
\begin{tabular}{c|c}
\hline
\multicolumn{2}{c}{\T  \bf Feature Aggregation Module $h$} \\ \hline 
\multicolumn{2}{c}{Input $1+n$ $D$-dim feature vectors}\\ 
\multicolumn{2}{c}{$\rightarrow$ output a $D$-dim feature vector}\\ \hline
Operation & Output Shape \\ \hline
Concat(\{$1+n$ features\}) & ($(1+n)\times D$,)\\\hline
Conv1D+ReLU & ($D$,)\\\hline
\end{tabular}
\vspace{-6pt}
\caption{Feature Aggregation Module. }
\end{table}

\begin{table}
\footnotesize
\centering
\label{table:supp-network-sdf}
\begin{tabular}{c|c}
\hline
\multicolumn{2}{c}{\T  \bf Signed Distance Function $f$} \\ \hline 
\multicolumn{2}{c}{Input a $3$-dim point and a $S$-dim feature vector $F_a$}\\ 
\multicolumn{2}{c}{$\rightarrow$ output a $1$-dim scalar}\\ \hline
Operation & Output Shape \\ \hline
Conv1D+ReLU & (64,)\\\hline
Conv1D+ReLU & (256,) \\\hline
Conv1D+ReLU & (512,), denoted as $P_f$ \\\hline
Concat($F_a$,$P_f$) & ($S$+512,)\\\hline
Conv1D+ReLU & (512,) \\\hline
Conv1D+ReLU & (256,) \\\hline
Conv1D+ReLU & (1,) \\\hline
\end{tabular}
\vspace{-6pt}
\caption{Signed Distance Function.}
\end{table}

The feature modulation network is attaching to the image encoder from which receives local image features and produces a new local image feature. Since the local image feature contains six sub-features, we devise six aggregation modules. Without losing generality, we set the dimension of the sub-feature to $D$.


The SDF regression network $f$ contains two MLPs, a global MLP receiving global image feature and a local MLP accepting local image feature. 
Without losing generality, we introduce a network with input a feature of dimension $S$. 
The 3D coordinate is firstly promoting to a $512$-dim point feature by three fully connected layers. Then the point feature $P_f$ and input feature $F_a$ are concatenated and passed to another three fully connected layers to generate the SDF value. The SDF value from global MLP and local MLP are added together as final output.

\section{More results} 

\textbf{More Quantitative Results.}
We includes more state-of-the-art methods on the single-image 3D reconstruction task.
All the methods, including 
AtlasNet~\cite{groueix2018papier}, Pixel2Mesh~\cite{wang2018pixel2mesh}, 
3DN~\cite{wang20193dn}, 
ImNet~\cite{chen2018implicit_decoder}, 
3DCNN~\cite{DISN}, 
OccNet~\cite{OccNet}, 
DISN~\cite{DISN} and Ladybird~\cite{xu2020ladybird}.

\begin{table*}[h!]
\footnotesize
\tabcolsep=0.095cm
\begin{center}
\begin{tabular}{l|l|ccccccccccccc|c}
\hline
Metrics & Methods & plane & bench & cabinet & car& chair& display & lamp & speaker& rifle & sofa& table & phone & watercraft & mean \\ \hline
\multirow{10}{*}{\textbf{CD}$\downarrow$} &AtlasNet & 5.98  & 6.98  & 13.76  & 17.04  & 13.21  & 7.18  & 38.21  & 15.96  & 4.59  & 8.29  & 18.08  & 6.35  & 15.85  & 13.19 \\
&Pixel2Mesh & 6.10  & 6.20  & 12.11  & 13.45  & 11.13  & \textbf{6.39}  & 31.41  & 14.52  & 4.51  & \textbf{6.54}  & 15.61  & 6.04  & 12.66  & 11.28 \\
&3DN & 6.75  & 7.96  & \textbf{8.34}  & 7.09  & 17.53  & 8.35 & \textbf{12.79}  & 17.28  & \textbf{3.26}  & 8.27  & 14.05  & 5.18  & 10.20  & 9.77 \\
&IMNET & 12.65  & 15.10  & 11.39  & 8.86  & 11.27  & 13.77  & 63.84  & 21.83  & 8.73  & 10.30  & 17.82  & 7.06  & 13.25  & 16.61 \\
&3DCNN & 10.47  & 10.94  & 10.40  & 5.26  & 11.15  & 11.78  & 35.97  & 17.97  & 6.80  & 9.76  & 13.35  & 6.30  & 9.80  & 12.30 \\
&OccNet & 7.70  & 6.43  & 9.36  & 5.26  & 7.67  & 7.54  & 26.46  & 17.30  & 4.86  & 6.72  & 10.57  & 7.17  & 9.09  & 9.70 \\
&DISN & 9.96 & 8.98 & 10.19 & 5.39 & 7.71 & 10.23 & 25.76 & 17.90 & 5.58 & 9.16 & 13.59 & 6.40 & 11.91 & 10.98 \\
&Ladybird & 5.85 & 6.12 & 9.10 & 5.13 & 7.08 & 8.23 & 21.46 & 14.75 & 5.53 & 6.78 & 9.97 & \textbf{5.06} & 6.71 & 8.60 \\
&Ours$_{cam}$ & \textbf{5.40} & \textbf{5.59} & 8.43 & \textbf{5.01} & \textbf{6.17} & 8.54 & 14.96 & \textbf{14.07} & 3.82 & 6.70 & \textbf{8.97} & 5.42 & \textbf{6.19} & \textbf{7.64}\\
&Ours & 3.27 & 3.38 & 6.88 & 3.93 & 4.40 & 5.40 & 6.77 & 8.48 & 1.58 & 4.38 & 6.49 & 4.02 & 4.01 & 4.85 \\
\hline
\T\T 
\multirow{10}{*}{\textbf{EMD}$\downarrow$} &AtlasNet &3.39  & 3.22  & 3.36  & 3.72 &  3.86  & 3.12  & 5.29  & 3.75  & 3.35  & 3.14  & 3.98  & 3.19  & 4.39  & 3.67 \\
&Pixel2Mesh & 2.98  & 2.58  & 3.44  & 3.43  & 3.52  & 2.92  & 5.15  & 3.56  & 3.04  & 2.70  & 3.52  & 2.66  & 3.94  & 3.34 \\
&3DN & 3.30  & 2.98  & 3.21 &  3.28  & 4.45  & 3.91  & 3.99  & 4.47  & 2.78  & 3.31  & 3.94  & 2.70  & 3.92  & 3.56 \\
&IMNET & 2.90  & 2.80  & 3.14  & 2.73  & 3.01  & 2.81  & 5.85  & 3.80  & 2.65  & 2.71  & 3.39  & 2.14  & 2.75  & 3.13 \\
&3DCNN & 3.36  & 2.90  & 3.06  & \textbf{2.52}  & 3.01  & 2.85  & 4.73  & 3.35  & 2.71  & 2.60  & 3.09  & 2.10  & 2.67  & 3.00 \\
&OccNet & 2.75  & 2.43  & 3.05  & 2.56  & 2.70  & 2.58  & 3.96  & 3.46  & 2.27  & \textbf{2.35}  & 2.83  & 2.27  & 2.57  & 2.75 \\
&DISN & 2.67  & 2.48  & 3.04  & 2.67  & 2.67  & 2.73 &  4.38 &  3.47 &  2.30  & 2.62  & 3.11  & 2.06 &  2.77  & 2.84 \\
&Ladybird & 2.48  & 2.29  & 3.03  & 2.65 &  2.60  & 2.61 & 4.20  & 3.32  & 2.22  & 2.42 &  2.82  & 2.06 &  2.46  & 2.71 \\
&Ours$_{cam}$ &\textbf{2.35} & \textbf{2.15} & \textbf{2.90} & 2.66 & \textbf{2.49} & \textbf{2.49} & \textbf{3.59} & \textbf{3.20} & \textbf{2.04} & 2.40 & \textbf{2.70} & \textbf{2.05} & \textbf{2.40}  & \textbf{2.57} \\
&Ours & 1.91 & 1.90 & 2.58 & 2.36 & 2.17 & 2.08 & 2.66 & 2.75 & 1.52 & 2.11 & 2.36 & 1.77 & 1.99 & 2.17 \\
\hline
\T\T
\multirow{10}{*}{\textbf{IOU}$\uparrow$} &AtlasNet & 39.2  & 34.2  & 20.7  & 22.0  & 25.7  & 36.4  & 21.3  & 23.2  & 45.3  & 27.9  & 23.3  & 42.5  & 28.1  & 30.0 \\
&Pixel2Mesh & 51.5  & 40.7  & 43.4  & 50.1  & 40.2  & 55.9  & 29.1  & 52.3  & 50.9  & 60.0  & 31.2  & 69.4 &  40.1  & 47.3\\
&3DN & 54.3  & 39.8  & 49.4  & 59.4  & 34.4  & 47.2  & 35.4  & 45.3  & 57.6  & 60.7  & 31.3  & 71.4  & 46.4  & 48.7 \\
&IMNET & 55.4  & 49.5  & 51.5  & 74.5  & 52.2  & 56.2  & 29.6  & 52.6  & 52.3  & 64.1  & 45.0  & 70.9  & 56.6  & 54.6 \\
&3DCNN & 50.6  & 44.3  & 52.3  & \textbf{76.9}  & 52.6  & 51.5  & 36.2  & 58.0  & 50.5  & 67.2  & 50.3  & 70.9  & 57.4  & 55.3\\
&OccNet & 54.7  & 45.2  & \textbf{73.2}  & 73.1  & 50.2  & 47.9  & 37.0  & \textbf{65.3}  & 45.8  & 67.1  & \textbf{50.6} &  70.9  & 52.1  & 56.4 \\
&DISN & 57.5  & 52.9  & 52.3  & 74.3  & 54.3 &  56.4  & 34.7 &  54.9  & 59.2  & 65.9  & 47.9 &  72.9 &  55.9  & 57.0 \\
&Ladybird & 60.0  & 53.4 &  50.8  & 74.5  & 55.3  & 57.8  & 36.2  & 55.6 &  61.0  & 68.5  & 48.6  & 73.6 &  \textbf{61.3}  & 58.2 \\
&Ours$_{cam}$ & \textbf{60.6} & \textbf{54.4} & 52.9 & 74.7 & \textbf{56.0} & \textbf{59.2} & \textbf{38.3} & 56.1 & \textbf{62.9} & \textbf{68.8} & 49.3 & \textbf{74.7} & 60.6 & \textbf{59.1} \\
&Ours & 68.2  & 63.1  & 61.4 &  80.7   & 66.8  &  67.9 & 55.9   &  65.0 &  75.0  & 75.2 &  62.6 & 81.0  & 68.9  & 68.6  \\
\hline
\end{tabular}
\caption{
Quantitative results on the ShapeNet Core dataset for various methods.}
\label{table:qr}
\end{center}
\vspace{-6pt}

\end{table*}

\begin{window}[8,c,{
\footnotesize
\tabcolsep=0.05cm
\begin{tabular}{l|c|c|c}
   &CD$\downarrow$  & EMD$\downarrow$ & IOU$\uparrow$ \\ \hline
Ours w/o s. p. & 15.02 & 3.20 & 58.2    \\ \hline
Ours     & 13.25  &  2.88   &  61.0     \\
\end{tabular}
},{}]
\textbf{More Ablation Study.}
Our network degenerates to DISN when the spatial pattern and its associated network modules are removed. To reduce the impact of network capacity, we train a variant with a network structure similar to Ours. 
To avoid introducing a point structure or geometric relation, we duplicate the input point to create a trivial pattern and feed it into the network. 
Numerical results on the Pix3D dataset prove the effectiveness of the spatial pattern. 
\end{window}